\documentclass{article}

\usepackage{microtype}
\usepackage{graphicx}
\usepackage{subfigure}
\usepackage{booktabs} % for professional tables

\usepackage{hyperref}

\usepackage[accepted]{icml2024}

\usepackage{amsmath}
\usepackage{amssymb}
\usepackage{mathtools}
\usepackage{amsthm}

\usepackage[capitalize,noabbrev]{cleveref}

\graphicspath{{imgs/}}
\DeclareGraphicsExtensions{.pdf,.jpeg,.png}
\usepackage{textcomp}
\def\BibTeX{{\rm B\kern-.05em{\sc i\kern-.025em b}\kern-.08em
    T\kern-.1667em\lower.7ex\hbox{E}\kern-.125emX}}
\usepackage{microtype}
\usepackage{graphicx}
\usepackage{booktabs} % for professional tables
\usepackage{amsmath}
\usepackage{amssymb}
\usepackage{mathrsfs}  
\usepackage{hyperref}
\usepackage{amsfonts}
\usepackage{booktabs}
\usepackage{siunitx}
\usepackage[switch]{lineno}   %% <- no mathlines option
\usepackage{amsmath}  %% <- after lineno
\usepackage{etoolbox} %% <- for \cspreto, \csappto
\newcommand*\linenomathpatch[1]{%
  \cspreto{#1}{\linenomath}%
  \cspreto{#1*}{\linenomath}%
  \csappto{end#1}{\endlinenomath}%
  \csappto{end#1*}{\endlinenomath}%
}
\linenomathpatch{equation}
\linenomathpatch{gather}
\linenomathpatch{multline}
\linenomathpatch{align}
\linenomathpatch{alignat}
\linenomathpatch{flalign}
\usepackage{enumitem}
\usepackage{indentfirst}
\usepackage{lineno}
\theoremstyle{plain}

\theoremstyle{definition}

\theoremstyle{remark}

\usepackage[textsize=tiny]{todonotes}

\icmltitlerunning{Reinforcement Learning for Solving  Stochastic Vehicle Routing Problem with Time Windows}

\begin{document}
\setlength{\abovedisplayskip}{2pt}
\setlength{\belowdisplayskip}{2pt}
\twocolumn[
\icmltitle{Reinforcement Learning for Solving \\ Stochastic Vehicle Routing Problem with Time Windows}



\begin{icmlauthorlist}
\icmlauthor{Zangir Iklassov}{1}
\icmlauthor{Ikboljon Sobirov}{1}
\icmlauthor{Ruben Solozabal}{1}
\icmlauthor{Martin Tak\'a\v{c}}{1}
\end{icmlauthorlist}
\icmlaffiliation{1}{Department of Machine Learning, MBZUAI, Abu-Dhabi, UAE}
\icmlcorrespondingauthor{Zangir Iklassov}{zangir.iklassov@mbzuai.ac.ae}

\icmlkeywords{Vehicle Routing Problem, Logistics Costs Optimization, Supply Chain, Reinforcement Learning}
\vskip 0.3in
]
\begin{abstract}
This paper introduces a reinforcement learning approach to optimize the Stochastic Vehicle Routing Problem with Time Windows (SVRP), focusing on reducing travel costs in goods delivery. We develop a novel SVRP formulation that accounts for uncertain travel costs and demands, alongside specific customer time windows. An attention-based neural network trained through reinforcement learning is employed to minimize routing costs. Our approach addresses a gap in SVRP research, which traditionally relies on heuristic methods, by leveraging machine learning. The model outperforms the Ant-Colony Optimization algorithm, achieving a 1.73\% reduction in travel costs. It uniquely integrates external information, demonstrating robustness in diverse environments, making it a valuable benchmark for future SVRP studies and industry application.
\end{abstract}

\section{Introduction}
\label{introduction}

% \begin{align}
% \text{min}\qquad &\textstyle{\sum}_{i, j \in N} c_{i j} x_{i j}+\mathscr{R}(x) \\
% d_{i} \sim F_{D} \\
% c_{ij} \sim F_{C} \\
% w \sim F_{W} \\
% W \sim U(-1, 1) \\
% d_{i}, c_{ij} = F(W) \\
% R(i) = 2r_{i}c_{0i}
% \end{align}

Reinforcement Learning (RL) aims to train machine learning (ML) models to optimize decision-making by maximizing reward outcomes while engaging with external environments. The remarkable achievements of RL in various domains like gaming \cite{silver2017mastering} and robotics \cite{Andrychowicz2020LearningDI} have ushered in a growing interest in its application to combinatorial optimization challenges \cite{Bello2017NeuralCO, Oroojlooyjadid2022ADQ, Tricks2018AND, iklassovSST23, iklassov2023reinforcement, IklassovMR023}. The golden standard for combinatorial optimization, which is famous for its NP-hard complexity, is to employ heuristic methods or specific solvers like Google Optimization Tools (a.k.a. Google OR-tools \yrcite{Google}). Although heuristic methods are known for their speed, they typically yield inferior solutions when compared to more sophisticated yet slower solvers. Reinforcement learning presents a viable alternative that mitigates these shortcomings. RL can be viewed as a technique that autonomously learns a set of heuristics from the provided data. As a result, it generally produces superior solutions compared to traditional heuristics, and it offers quicker inference times than solver-based approaches.

\begin{figure}%[!ht]
\centering
\includegraphics[width=0.9\linewidth]{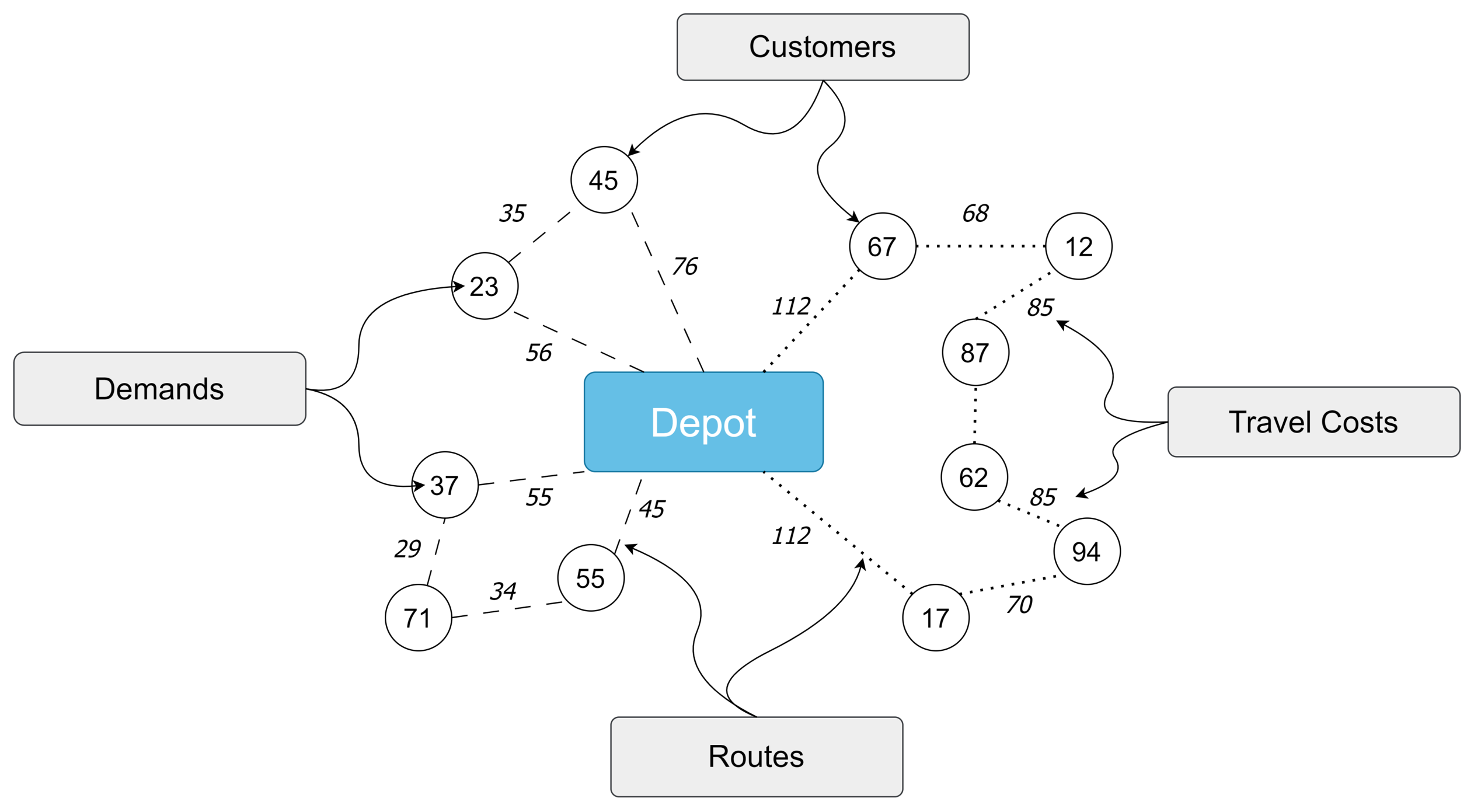}
\caption{This Vehicle Routing Problem instance involves a graph with customer nodes and a central depot, all positioned within a two-dimensional coordinate system, where each customer node has a demand for goods and links between nodes have associated travel costs. Vehicles start from the depot and visit these customer nodes sequentially. Our formulation, tailored for industrial use, includes stochastic demands and undisclosed travel costs, as well as time windows for goods delivery. The decision-making agent operates in this stochastic environment, aiming to learn and adapt strategies to reduce expected travel costs.}
\label{fig:example}
\end{figure}

{\bf Vehicle Routing Problem} (VRP) is a fundamental problem in combinatorial optimization, in which the goal is to identify the most effective routing scheme for a number of vehicles with restricted carrying capacity; these vehicles are assigned to deliver objects or services to different customers, with an objective to reduce the cumulative travel cost (refer to Figure \ref{fig:example}). In this context, Nazari et al. \yrcite{nazari2018reinforcement} pioneered a methodology based on an RL model capable of addressing the VRP via the use of a pointer network. 

{\bf Stochastic Vehicle Routing Problem} with time windows (SVRP) is a complex formulation of the VRP, characterized by its unknown parameters that are only revealed after routes are completed. These uncertain parameters make the SVRP challenging problem compared to the deterministic VRP. At the same time SVRP reflects real-world scenarios frequently faced in the industrial sector, and therefore, solutions to the SVRP are more desirable \cite{inbook, GENDREAU19963}. Traditional practices to tackle SVRP have been examined in various studies, but the application of deep RL to this formulation of the VRP remains unexplored.

{\bf Motivation.} Our primary motivation for investigating the SVRP lies in the lack of research in general SVRP employing RL methodologies, although it may provide substantial advantages for businesses in transportation and logistics. They include, but are not limited to, lowering carbon emissions and reducing operational costs. This research focuses on improving the applicability of SVRP frameworks \cite{iklassovSST23, iklassov2023reinforcement} in supply-chain contexts by integrating time windows into SVRP, addressing two key stochastic elements: demand and travel costs. We perform a series of experiments to evaluate how our proposed model responds to various configurations of the SVRP environment, commonly investigated in traditional studies. Our specific {\bf contributions} to this field are summarized as follows:
\begin{itemize}
    \item We propose \textbf{the first RL-based model for SVRP} that integrates time windows, stochastic customer demands, and travel costs, pivotal elements in practical logistics scenarios. Through a comparative analysis against contemporary SVRP methods, our RL model showcases superior performance, manifesting a 1.73\% decrease in travel costs in relation to the most proficient classical model \cite{Goel2019VehicleRP}.
    \item We conduct comprehensive experiments to evaluate the performance of our model across a spectrum of environmental configurations. These configurations encompass diverse inference methodologies, delivery modalities, levels of stochasticity, fleet and customer magnitudes. Our observations indicate that \textbf{the RL agent exhibits robustness across distinct environmental settings}, a crucial attribute for SVRP given its propensity for varied formulations in real-world scenarios.
    \item We demonstrate that \textbf{the model is capable to leverage external information}, facilitating the derivation of improved routing strategies. This aspect holds particular significance in the SVRP industry, an aspect that has hitherto received limited attention in scholarly discourse. We posit that our model constitutes a valuable baseline for both research and industrial applications, providing a foundation for prospective investigations. The complete framework, inclusive of all pertinent source codes, is accessible online.\footnote{\url{https://github.com/Zangir/SVRP}}
\end{itemize}

\section{Related Work}
\label{related_work}

{\bf VRP.} RL has seen a prevalent use in deterministic VRP in recent years. Nazari et al. \yrcite{nazari2018reinforcement} can be considered the pioneering work in applying RL for deterministic VRP using a pointer network-backed framework. Their approach performed superior to standard heuristic and meta-heuristic baselines. Lu et al. \yrcite{Lu2020ALI} studied a meta-heuristic agent solution based on RL that coordinates a set of rules to search for an optimal VRP solution. While their method reaches the state-of-the-art (SOTA) performance with the lowest cost for VRP instances with 50-100 nodes, the solution requires domain knowledge as an input to create a set of possible heuristics. To address this issue, a new RL pipeline was suggested by Li et al. \yrcite{Wang2021AlphaTLT} who achieved superior performance for bigger problem instances including 500, 1000, and 2000 customers.
\\[2pt]
{\bf SVRP.} With numerous solutions in deterministic VPR using RL, incorporating these methods into SVRP cases continues to pose difficulties. Toth and Vigo \yrcite{Toth2014VehicleRP} summarize the solutions for SVPR based solely on the classical algorithms, categorizing them into branch-and-bound methods (including branch-and-cut \cite{Gauvin2014ABA}, branch-and-price \cite{Christiansen2007ABA, FUKASAWA202311} and integer L-shaped \cite{Laporte1993TheIL, DELAVEGA2023676}), heuristic methods \cite{Laporte2009FiftyYO} (including LKH3 \cite{Helsgaun2017AnEO} and Clarke-Wright \cite{Pichpibul2013AHA}), and metaheuristic methods \cite{Dantzig1959TheTD} (including tabu search \cite{Li2020AnIT, su15021741} and ant-colony optimization \cite{Goel2019VehicleRP}). Pichpibul and Kawtummachai \yrcite{Pichpibul2013AHA} proposed the Clarke-Write algorithm that addresses both deterministic and stochastic VRP challenges using a simplified heuristic model. Their methodology has become a standard baseline in subsequent literature. Following this, the focus shifted increasingly towards metaheuristic-based methods for SVRP. Among these, Ant Colony Optimization (ACO) has emerged as a dominant technique, achieving new benchmarks in solution quality \cite{Goel2019VehicleRP}. The application of reinforcement learning in the field of SVRP is emerging as a significant area of research. Several studies have proposed various SVRP formulations and corresponding RL-based solutions \cite{ZHANG2023596, ZHOU2023109443, HILDEBRANDT2023106071, JIN2023}. However, these formulations focus on particular VRP scenarios and do not fully address the broader scope of general SVRP, which encompasses all key sources of stochasticity in VRP, such as demand, travel costs, and time windows. Additionally, they lack the integration of external information that could significantly improve their applicability in industrial settings.
\\[2pt]
{\bf Research Gap.} Commercial enterprises frequently incorporate external information, such as meteorological data, into their estimations of stochastic variables \cite{Bahaabadi2021, Bomboi2021OnTS}. This practice potentially enables agents to derive enhanced routing strategies. However, a gap exists in research with respect to this particular direction of investigation. Consequently, we introduce external variables into our model to evaluate the model's capacity to acquire efficacious policies through new information.

\section{Method}
\label{method}

{\bf Formulation.} In the SVRP, there exists inherent uncertainty as the actual values of customer demands and travel costs remain undisclosed to the agent. In this scenario, the agent possesses probabilistic or expected values for both the customer demands and travel costs, but upon reaching a particular customer, there is a distinct possibility that either the demand or the travel costs associated with that customer may deviate from their expected values. In addressing this stochastic variation, industries often employ implicit variables to manage the inherent uncertainty. For instance, companies may utilize dynamic information, such as weather forecasts for the following day, to adapt their expectations regarding travel costs. This adaptive approach allows companies to make informed decisions and enhance the resilience of their vehicle routing strategies in the face of uncertain and dynamic factors.
\begin{table} % title of Table
\centering % used for centering table
\caption{Notations used throughout the paper.}
\begin{tabular}{c@{\hskip 0.2in}l}% centered columns (4 columns)
\\
\toprule %inserts double horizontal lines
\textbf{Notation} & \textbf{Description} \\ %[0.5ex] % inserts table
%heading
\midrule % inserts single horizontal line
% & \multicolumn{2}{c}{Four Rooms }\\
$N$	& set of customers and depot \\
\hline
$C$	& customers set \\
\hline
$c_{ij}$	& travel cost \\
& between nodes $ij$\\
\hline
$d_{i}$	& demand of customer $i$\\
\hline
$z_{i}$	& type of customer $i$\\
\hline
$H_{i}$	& time windows of customer $i$\\
\hline
$K$	& set of vehicles \\
\hline
$q$	& current load of a vehicle \\
\hline
$p$	& maximum load of each vehicle \\
\hline
$q$	& current position of a vehicle \\
\hline
$w$	& external variable affecting \\
& demand and travel cost \\
\hline
$\theta$	& stochastic variable \\
& demand or travel cost \\
\hline
$t$	& current time in hours \\
\hline
$r$	& recourse action \\
& to return to the depot \\
\hline
$x_{i j}$	& binary action variable showing \\
& whether $(i, j)$ is used in the route \\
\bottomrule
\end{tabular}
\label{tab:variables}
\vskip -18pt
\end{table}
\\[2pt]
{\bf Notations.} To formalize the problem, we establish the following mathematical notation (Table~\ref{tab:variables}). We define a set $N$, encompassing both the customers and the depot. For each customer $i$ within this set, we introduce a variable $d_{i}$ to denote their specific demand. Additionally, we define a variable $q_{k}$ for each vehicle $k$ within the fleet $K$, representing the current cargo load of vehicle $k$. The maximum load capacity is indicated by $Q$. The cargo load varies over time, reflecting the amount of goods that vehicle $k$ is carrying at any specific moment. We introduce a cost variable $c_{ij}$, which characterizes the cost associated with traversing the arc connecting node $i$ to node $j$. It quantifies the time incurred in moving from one node to another within the problem instance. We introduce a binary decision variable $x_{ij}$, which assumes a value of one when vehicle $k$ traverses the arc between nodes $i$ and $j$ as part of its route and takes on a value of zero otherwise. The binary nature of this variable effectively captures whether a given arc is utilized within the vehicle's route. The routing process is developed iteratively, with the location of each vehicle being updated at every time step $t$. Inherently, our problem structure permits travel between any pair of nodes in the graph.
\\[2pt]
{\bf Failure Situation.} In the context of the SVRP, scenarios may recall when a vehicle arrives at a customer location and finds that the customer's demand exceeds both the initially estimated value and the current load of the vehicle. In such instances, the vehicle is compelled to return to the depot to replenish its full capacity ($q_k = Q$), before returning to address the demand of this particular customer. This necessitates incurring an additional cost associated with the so-called failure situation, which entail substantial expenses within industrial contexts and warrant the implementation of improved routing strategies. To address this challenge, one potential solution involves anticipating higher demand from a specific distant customer and strategically refilling the vehicle in advance while in proximity to the depot. However, failure situations may still be presented in the solution. To catch them, we introduce the binary variable $r_i$ representing the enforced decision to undertake the action of returning to the depot from customer $i$ for refilling. Specifically, if $r_i = 1$, it indicates the necessity of the vehicle to proceed to the depot from customer $i$ for refilling purposes. 
\\[2pt]
{\bf Objective.} The main goal of the agent is to minimize the total cost by selecting appropriate values for the binary variables $x_{ij}$. This involves reducing the combined expenses of all routes and costs arising from failure scenarios, while simultaneously ensuring that customer demands within the network are adequately met, i.e.:
\begin{align}
\text{minimize} \quad &\textstyle{\sum}_{i, j \in C} c_{ij} x_{ij}+2\textstyle{\sum}_{i \in C} r_{i}c_{0i}. \nonumber 
\end{align}
{\bf External Variables.} We introduce a set of external variables $W$, with their values known to the agent. These variables exert a nonlinear influence on the realizations of stochastic variables, specifically demands and travel costs. In industrial applications, logistic companies often leverage such external variables to refine their estimations of stochastic variables. For instance, for a company specializing in ice cream sales, knowledge of weather conditions enables more accurate estimations of ice cream demand for the following day. In this context, we define a set of external variables, each conforming to a certain distribution.
\\[2pt]
{\bf Stochastic Variables.} The agent lacks knowledge regarding the distributions and exact values of stochastic variables. Nevertheless, the agent may possess estimations for the realizations of these variables, along with the values of external variables. The external variables exert an impact on the stochastic demand and travel cost according to the following relationship:
\begin{align}
\theta_{i} = \bar{\theta}_{i} + \textstyle{\sum}_{m} \textstyle{\sum}_{n} \alpha_{imn} w_{m} w_{n} +\epsilon_i \text{,   }\\
\theta_{i} \in \{ d_{i}, c_{i0}, ..., c_{iN}\},
\nonumber
\end{align}
where $\theta_i$ serves as a stochastic variable, representing either the demand associated with customer $i$ or the travel cost from customer $i$ to other nodes. The value of $\theta_i$ is a composite of a fixed term, denoted as $\bar{\theta}_i$, the cumulative effect of the weighted interaction between all pairs $w_m \in W$ and $w_n \in W$, and the random noise denoted as $\epsilon_i$. Regarding travel costs, the fixed term is defined as the Euclidean distance between two nodes divided by a constant speed. The travel cost, in this context, is intended to signify the time required to travel from one node to another. It is noteworthy that the agent lacks explicit knowledge concerning the distribution of variables within the set $W$ or their influence on $\theta_i$. However, the agent is endowed with access to the observed realizations of variables in $W$, affording the opportunity to implicitly learn the nonlinear mapping from $W$ to stochastic variables.
\\[2pt]
{\bf Time Windows.} We define a set of customer types to clusterize the customers (e.g., office workers vs housekeepers). In the problem-solving process, we maintain a record of the current environmental time denoted as $t$, measured in hours, with travel costs also quantified in hours. Additionally, we establish time windows and handout times for each customer, stratified by customer type. Each day is characterized by $h_{max}$ hours. To encapsulate the temporal availability of each customer, we introduce a binary vector $H_i$ for every customer $i$. This vector conveys the customer's availability at each hour of the day. Specifically, $H_i^t = 1$ signifies the availability of customer $i$ at hour $t$, indicating the feasibility of delivering the product during that period. Distinct probabilities dependent on customer types govern the availability of customers during specific hours. For each problem instance, we generate the $H_i$ vector based on this probability distribution. The agent possesses information regarding the customer type, current time, and realization of the $H_i$ vector for each customer.

\subsection{Baselines}
{\bf Clarke-Wright heuristic.} The literature on SVRP comprises diverse baseline approaches, among which CW heuristic  is notable \cite{Pichpibul2013AHA}. This method, established on the concept of \textit{savings}, evaluates the reduction in total travel cost caused by combining two customer nodes into a single route. The \textit{saving} for any two nodes $i$ and $j$ is calculated as $saving_{ij}=\mathbb{E} [c_{0i}]+\mathbb{E} [c_{j0}]-\mathbb{E} [c_{ij}], $ where $c_{0i}$ and $c_{0j}$ are the costs of traversing from the depot to consumers $i$ and $j$ respectively, and $c_{ij}$ is the cost between these customers. These \textit{savings} are then compiled into a list for all customer pairs, sorted from highest to lowest order. The process of merging the routes starts with the selection of the pair with the largest \textit{saving}, combining them into one route if their combined expected demand $\mathbb{E} [d_{i}] + \mathbb{E} [d_{j}]$ does not exceed the vehicle maximum capacity $Q$. The approach continues until no other efficient merges can be constructed from the \textit{savings} list. 
\\[2pt]
{\bf Tabu Search (TS).} The TS algorithm
\cite{Li2020AnIT} is another prominent optimization technique for solving the SVRP that relies on metaheuristics. The TS method begins with a randomly created potential solution, and its total cost is evaluated. It involves using a succession of neighborhood heuristic operations to this initial solution, generating a new log of feasible solutions over a specified number of iterations, designated as $k_{Tabu, max}$. The method finishes by determining the optimal candidate from all the potential solutions it has examined.
\\[2pt]
{\bf Ant-Colony Optimization (ACO).} ACO is a metaheuristic method \cite{Goel2019VehicleRP} inspired by the foraging behavior of ants in nature. In this approach, agents searching for solutions operate like ants, traversing the search space to investigate candidate routes. Once an agent pinpoints a potential solution, it deposits a pheromone trail on the arcs forming that route. The likelihood of an arc being an optimistic choice is affected by the intensity of the pheromone left by the agent, navigating the decisions of succeeding agents. In essence, a more intense pheromone concentration on an arc denotes its better quality as a candidate route. This pheromone evaporates over time, mirroring the ant behavior in nature. The pursuit of candidate solutions continues as the ants examine further arcs.

\subsection{Routing Policy}

\textbf{State}. Within the context of the RL framework, the state representation encompasses a multifaceted set of variables, including the current customer demands, travel costs, and the present positions and cargo loads of the vehicles. Our formulation delineates the environment through a state representation $(I,L)$, wherein $I$ encapsulates information pertaining to each customer, and $L$ encapsulates information pertinent to each vehicle.
\begin{align}
 I &\doteq \{(W_{i}, d_{i}, C_{i}, H_{i}, Z_{i}), i \in N\} \text{,   }\ \\
 L &\doteq \{(q_{k}, p_{k}, t_{k}), k \in K\}, \nonumber
\end{align}
where $W_{i}$ is the vector of external variables, $d_{i}$ represents the realization of the demand for customer $i$, $C_{i}$ signifies the vector of realizations of travel costs from customer $i$ to other nodes. $H_{i}$ denotes the time windows vector of customer $i$, and $Z_{i}$ is a binary vector indicating the type of customer, $q_{k}$ represents the load of vehicle $k$, $p$ denotes its position, and $t$ signifies the current time. This constitutes the environment time, which is similar across all vehicles. It is crucial to highlight that the values of customer demand, vehicle capacity, and vehicle position are subject to change as time progresses.
\\[2pt]
{\bf Action.} The actions undertaken within this framework pertain to the manipulation of the vehicle's spatial positioning. At every discrete time step $t$, the agent makes decisions by selecting an action $a^{t}$. Here, the action pertains to determining the subsequent position $p^{t+1}_k$ of each vehicle $k$. The decision-making process is dynamic and is orchestrated by a policy $a^{t} \sim \pi(\cdot| I^t, L^t)$. Indicating that the actions at time $t$ are drawn from a probability distribution governed by the policy $\pi$, conditioned on the current state  ($I^t, L^t)$. Succinctly, the policy dictates the likelihood of selecting the next node for each vehicle based on the prevailing state information.
\\[2pt]
{\bf Transitions.} The transition function governing state dynamics encompasses alterations in vehicle position, cargo load, and customer demand. When a vehicle arrives to customer $i$ at time $t$, it mitigates the demand for time $t + 1$ by an extent equivalent to the current cargo load carried by vehicle $k$. It is imperative to underscore that the minimum allowable demand is set at zero. Simultaneously, the cargo load of vehicle $k$ diminishes by the magnitude of the current demand associated with customer $i$. When the cargo load of vehicle $k$ reaches zero, it is incumbent upon the vehicle to return to the depot for replenishment, thereby restoring its cargo-carrying capacity as follows:
\begin{align}
d^{t+1}_i &= \max\left\{0,d^{t}_i-q^t_k\right\} \text{, }\ \\
d^{t+1}_j &= d^{t}_j, \text{ for }j\neq i, \text{ and }\  \\
q^{t+1}_k &= \max\left\{0,q^t_k-d^{t}_i\right\}. \label{eq:vrp:dem-load1}
\end{align}
\textbf{Objective.} The objective of the RL model is the minimization of the anticipated travel cost, entailing an optimization process that seeks to enhance the efficiency and cost-effectiveness of vehicle routing. The objective of the agent is to formulate an optimal policy $\pi$ with parameters $\Theta$ through the utilization of the Reinforce algorithm \cite{williams1992simple}. This algorithm is designed to minimize the cumulative expected cost as follows:
\begin{align}
    \mathcal{J}^{\pi}(\Theta) = \mathbb{E} [\textstyle{\sum}_t^T C_t(I^t, L^t, a^t)]. \nonumber
\end{align}
The cost function $C_t(I^t, L^t, a^t)$ is defined as the summation of all traversal costs subsequent to the execution of action $a^t$. Mathematically, this is expressed as $\textstyle{\sum}_{k=1}^{K} c_{i_{k}j_{k}}$. In instances where vehicle $k$ necessitates recourse action due to a failure, an additional recourse cost of $2c_{0j_{k}}$ is incurred. The objective function is aggregated across $T$ time steps, and the gradient of the objective function \cite{sutton1999policy} is computed in the following manner:
\begin{align}
\nabla_\theta{\hat{\mathcal{J}}^{\pi}}(\Theta) \approx \nonumber \textstyle{\sum}_{t=1}^{T} ( &(C(I^t, L^t, a^t) - b_\phi(I^t, L^t)) \cdot \\  &\nabla_\Theta \log{\pi_\Theta(a^t | I^t, L^t)}). \nonumber
\end{align}
In this context, ${b_\phi}(I^t, L^t)$ represents the baseline function characterized by parameters $\phi$, and these parameters are optimized through training to minimize
\begin{align}
    L(\phi) =  \textstyle{\sum}_{t=1}^{T} ||b_\phi(I^t, L^t) - C(I^t, L^t, a^t) ||^2. \nonumber
\end{align}
{\bf Inference.} For the purposes of inference, we will employ three distinct inference strategies commonly utilized in reinforcement learning literature. The first strategy is greedy inference, wherein we choose only one action at each time step $t$ with the highest probability determined by the policy function \cite{bello2016neural, kool2018attention}. The second strategy is sampling inference, involving the selection of $n_s$ actions at each time step based on their respective probability distributions. The third strategy is beam search inference, where $n_b$ trajectories of actions with the highest probability distributions are chosen \cite{joshi2019efficient, wang2021dynamic}.
\\[2pt]
{\bf Apriori vs. Reoptimization.} Two distinct approaches, namely apriori and reoptimization, are utilized for inference. In the apriori approach, the agent devises all routes solely based on estimates of stochastic variables, with realizations being disclosed after problem resolution. In the reoptimization approach, the agent discloses stochastic variable realizations during the creation of the solution. For instance, the agent discloses the actual demand of specific customers upon reaching them on the route. Both approaches find applicability in industrial scenarios and, therefore, merit evaluation.

\subsection{Environment Settings}

In this section, we explore various approaches for setting up the environment. We deem it crucial to address these approaches, as each setting can find its relevance in diverse industrial contexts.
\\[2pt]
{\bf Signal Ratio.} When constructing stochastic variables, we employ the signal ratio, a metric that gauges the impact of each component on the variable itself. Consequently, three distinct signal ratios are computed for the fixed term $A_{i}$, external variables term $B_{i}$, and noise $\Gamma_{i}$. Each signal ratio quantifies the proportion of the squared value of the stochastic variable contributed by the squared value of its respective terms. The signal ratio is a scalar measure ranging between zero and one computed as follows:
\begin{align*}
A_{i} = \tfrac{\bar{\theta_{i}^{2}}}{T_i} \quad
B_{i} = \tfrac{\mathbb{E}\left[(\sum_{m}\sum_{n}\alpha_{imn}w_{m}w_{n})^{2}\right]}{T_i},  \quad
\Gamma_{i} = \tfrac{\mathbb{E}\left[\epsilon_{i}^{2}\right]}{T_i}, \end{align*}
\begin{align}
T_{i} = \bar{\theta_{i}^{2}} + \mathbb{E} [(\textstyle{\sum}_{m}\textstyle{\sum}_{n}\alpha_{imn}w_{m}w_{n})^{2} ] + \mathbb{E}\ [\epsilon_{i}^{2} ]. \nonumber
\end{align}
{\bf Fill rate.} We additionally quantify the fill rate, defined as the ratio of the maximum capacity $Q$ of each vehicle to the aggregated expected demand from all customers, $\Phi = \frac{Q}{\mathbb{E}[d_i]}.$
\\[2pt]
{\bf Variable Estimates.} Prior to disclosing the actual realizations of stochastic variables, agents are equipped with initial estimates of these variables using fixed terms. The agent employs these fixed terms as initial approximations for the stochastic variables and updates them subsequent to the revelation of their true realizations.
\\[2pt]
{\bf Customer Positions.} Two approaches are considered regarding customer positions: flexible and fixed. In the flexible approach, it is assumed that for each problem instance, the locations of customers on the map can vary. Conversely, in the fixed approach, the locations of customers remain constant, although stochastic travel demands and travel costs can still be incorporated. 
\\[2pt]
{\bf Delivery Types.} Two delivery approaches can be considered. The first approach is partial delivery, allowing for the fulfillment of demand in part initially, with the remaining fulfillment scheduled for subsequent time steps. The second approach is exclusive to full delivery, restricting the fulfillment of demand to occur only within the same time step.

\section{Architecture}

\begin{figure}%[!ht]
\centering
\includegraphics[width=0.9\linewidth]{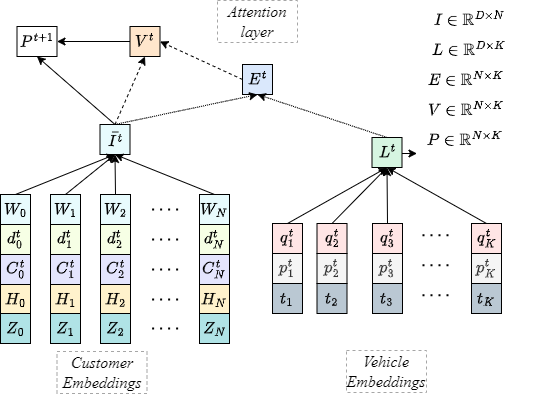}
\caption{\textbf{Model Architecture.} The lower segment depicts the input structure of the model. The model produces embeddings for two input types: Customer and Vehicle. These embeddings are then merged through an Attention Layer, yielding probabilities assigned to the nodes. These probabilities indicate the likelihood of each node being the subsequent position for each vehicle. $0$ node stands for the depot that has zero demand and is available all the time. Ultimately, the probabilities undergo a masking process to exclude customers within the route that have already been satisfied.
}
\label{fig:Network}
\vskip -18pt
\end{figure}

{\bf Inputs.} The architecture comprises three primary components (Figure \ref{fig:Network}). The first pertains to customer-related information input, encompassing details such as external variables, demand estimation, travel cost estimations, customer type, and time windows. The second input pertains to vehicle information, involving the load, position, and time $t$ for each vehicle. Time is encoded as a one-hot vector, with 1 representing the current hour. 
\\[2pt]
{\bf Embeddings.} Customer inputs undergo a one-dimensional convolutional layer to generate D-dimensional embeddings for each customer. Vehicle inputs undergo an LSTM layer, yielding same D size embeddings for each vehicle. This layer is designed to retain information about the routes traversed by each vehicle. Although previous routes theoretically do not impact future costs, in practice, the memory layer has shown a positive impact on model performance. Vehicle and customer embeddings pass through an attention layer, producing a matrix $P^{t+1}$ with a number of rows equal to the number of nodes and a number of columns equal to the number of vehicles. 
\\[2pt]
{\bf Action Selection Strategy.} The softmax function is applied to the columns of the matrix $P^{t+1}$, generating visitation probabilities for each vehicle to each node, encompassing both customer locations and the depot. These probabilities are then adjusted through a masking technique, effectively setting probabilities to zero for customers whose demands have already been met. The action for each vehicle is selected based on the highest probability. Under this approach, there's a possibility that two vehicles may visit the same customer, particularly when a customer's demand exceeds the carrying capacity of a single vehicle.

\section{Experiments}

{\bf Dataset.} During the training process for each batch, a two-dimensional map with coordinates ranging from zero to one is simulated. The depot is positioned at the center of this map. Uniform distribution is employed to generate values for three external variables, which are then utilized to simulate demands and travel costs. The simulation is conducted in a manner that adheres to the specified default signal ratio values for each experiment. Employing the same procedure, a test dataset comprising 1000 problem instances is generated and consistently used across all experiments. The model is trained and tested across scenarios involving 10, 20, 50, and 100 customers, employing one to five vehicles in multi-vehicle settings.
\\[2pt]
{\bf Default Environment.} The default environmental configuration involved employing beam search inference with $n_b = 3$, fixed-term stochastic variable estimations, fixed customer positions, partial type delivery, a fill rate of $\Phi = 0.5$, a single vehicle, and signal ratios $A, B, \Gamma$ set to $0.6, 0.2, 0.2$, respectively.

\subsection{Results}

{\bf Baselines.} To set up our baseline methods, we adopted a combination of cutting-edge models and widely used baselines from the general SVRP literature \cite{Pichpibul2013AHA, Li2020AnIT, Goel2019VehicleRP}. We opted not to include current RL approaches to SVRP, as they are tailored for particular instances of SVRP and fail to encompass all essential sources of stochasticity, as well as external variables. 

Subsequently, we assessed the performance of our proposed model against these baselines, as delineated in Table \ref{baselines_table}. The results demonstrate that our model, employing beam search inference, consistently achieves an average reduction of 1.73\% in travel costs across varying problem sizes compared to the ACO model. To ensure a fair comparison, we maintained a consistent default environment setting with apriori inference across all models in this evaluation.
\begin{table}
\caption{The reults obtained from the Baseline and RL (with beam search) models on the SVRP test dataset.}
\label{baselines_table}
\centering
\begin{small}
\begin{sc}
\scalebox{0.8}{
\begin{tabular}{crrrrr}
\\
\toprule
    Baselines &    10 &    20 &     50 &    100 \\
\midrule
    CW & 6.88 & 11.70 & 23.19 & 54.99 \\ 
    Tabu & 6.66 & 11.31 & 22.42 & 53.17 \\ 
    ACO  & 6.48 & 11.02 & 21.84 & 51.81 \\ 
    RL & \textbf{6.37} & \textbf{10.83} & \textbf{21.46} & \textbf{50.90} \\ 
 
\bottomrule
\end{tabular}
}
\end{sc}
\end{small}
% \vskip -8pt
\caption{The effectiveness of inference strategies on the SVRP test dataset, with reported outcomes for both the a priori (Apr.) and reoptimization (Reopt.) configurations.}
\label{select_strats_table}
\centering
\begin{small}
\begin{sc}
\scalebox{0.63}{
\begin{tabular}{c cc cc cc cc}
\\
\toprule

 & \multicolumn{2}{c}{{{10}}}
 & \multicolumn{2}{c}{{{20}}}
 & \multicolumn{2}{c}{{{50}}}
 & \multicolumn{2}{c}{{{100}}}\\

 \cmidrule[0.4pt](lr{0.125em}){2-3}%
 \cmidrule[0.4pt](lr{0.125em}){4-5}%
 \cmidrule[0.4pt](lr{0.125em}){6-7}%
 \cmidrule[0.4pt](lr{0.125em}){8-9}%
 
 Inference
 & Apr.
 & Reopt.
 & Apr.
 & Reopt.
 & Apr.
 & Reopt.
 & Apr.
 & Reopt.\\
 
\midrule
     Greedy & 7.22 & 7.01 & 12.28 & 11.91 & 24.34 & 23.61 & 57.72 & 55.99 \\ 
     Sampling & 6.71 & 6.51 & 11.41 & 11.07 & 22.61 & 21.93 & 53.63 & 52.02 \\  
     Beam Search  & \textbf{6.37} & \textbf{6.18} & \textbf{10.83} & \textbf{10.50} & \textbf{21.46} & \textbf{20.82} & \textbf{50.90} & \textbf{49.37} \\ 

\bottomrule
\end{tabular}
}
\end{sc}
\end{small}
\vskip -18pt
\end{table}
\\[2pt]
{\bf Robustness to Environment Changes.} To assess the model's robustness across various formulations of the SVRP environment, we conducted evaluations under diverse settings. Our experimentation spanned different inference strategies, varying levels of environmental stochasticity, signal ratios, fill rates, and customer positions.

In terms of inference strategies, we compared the performance of greedy inference against random sampling inference (with a width set to 16) and beam search inference (with a width set to three). The outcomes of these experiments, detailed in Table~\ref{select_strats_table}, illustrate that random sampling outperforms greedy inference, while beam search exhibits superior performance compared to both. Consequently, we adopted beam search inference for all subsequent experiments.
In the subsequent experiment, we evaluated the model's performance across different levels of stochasticity in the environment. The first experiment involved a fully deterministic setting, where the values of demand and travel costs were solely determined by their constant components. In the second experiment, we introduced stochastic demand with default ratio values $A, B, \Gamma = 0.6, 0.2, 0.2$. The third experiment focused on an environment featuring only stochastic travel costs, again with default ratio values. Subsequently, the model was tested in an environment with both stochastic demands and travel costs. Finally, we introduced time windows to the environment. The outcomes of these experiments are presented in Table \ref{source_table}. It is evident that with each added layer of complexity, the average travel costs increase. However, the model consistently demonstrates its capacity to adapt to varying problem conditions and optimize routes accordingly.

We fixed the environment, incorporating stochastic demands, travel costs, and time windows while manipulating the values of the signal ratios. The outcomes of this experiment are displayed in Table \ref{signal_ratio_table}. The results illustrate that as the impact of the noise term intensifies, the stochastic variable becomes more unpredictable, resulting in higher average travel costs. Simultaneously, it is evident that the effect of external variables is less erratic, indicating that the model successfully captures some of the influence of external variables on the stochastic variables. Consequently, the optimal outcome is achieved in a scenario where there is an effect solely due to external variables, without any accompanying noise.
\begin{table}
\caption{The results of different stochasticity source settings on the SVRP test dataset. D. - demand, Tr.C. - Travel Cost, T.W. - Time Windows}
\centering
\label{source_table}
    \scalebox{0.67}{
        \begin{tabular}{l cc cc cc cc} 
\\
\toprule

 & \multicolumn{2}{c}{{{10}}}
 & \multicolumn{2}{c}{{{20}}}
 & \multicolumn{2}{c}{{{50}}}
 & \multicolumn{2}{c}{{{100}}}\\

 \cmidrule[0.4pt](lr{0.125em}){2-3}%
 \cmidrule[0.4pt](lr{0.125em}){4-5}%
 \cmidrule[0.4pt](lr{0.125em}){6-7}%
 \cmidrule[0.4pt](lr{0.125em}){8-9}%
 
 Source
 & Apr.
 & Reopt.
 & Apr.
 & Reopt.
 & Apr.
 & Reopt.
 & Apr.
 & Reopt.\\
 
\midrule
     No Stoch. & 2.88 & 2.79 & 5.88 & 5.71 & 12.26 & 11.90 & 26.93 & 26.12 \\ 
     D. & 3.13 & 3.04 & 6.39 & 6.20 & 13.33 & 12.93 & 29.27 & 28.39 \\ 
     Tr.C. & 3.38 & 3.28 & 6.90 & 6.70 & 14.40 & 13.96 & 31.61 & 30.67 \\ 
     D./Tr.C. & 3.51 & 3.40 &  7.16 & 6.95 & 14.93 & 14.48 & 32.78 & 31.80 \\ 
     D./Tr.C./T.W. & 6.37 & 6.18 & 10.83 & 10.50 & 21.46 & 20.82 & 50.90 & 49.37 \\ 

 \bottomrule
        \end{tabular}
    }
\centering
\caption{The results derived from different signal ratios on the SVRP test dataset.}
\label{signal_ratio_table}
    \scalebox{0.63}{
        \begin{tabular}{c cc cc cc cc} 
\\
\toprule

 & \multicolumn{2}{c}{{{10}}}
 & \multicolumn{2}{c}{{{20}}}
 & \multicolumn{2}{c}{{{50}}}
 & \multicolumn{2}{c}{{{100}}}\\

 \cmidrule[0.4pt](lr{0.125em}){2-3}%
 \cmidrule[0.4pt](lr{0.125em}){4-5}%
 \cmidrule[0.4pt](lr{0.125em}){6-7}%
 \cmidrule[0.4pt](lr{0.125em}){8-9}%
 
 $\mathrm{A}/\mathrm{B}/\Gamma$ 
 & Apr.
 & Reopt.
 & Apr.
 & Reopt.
 & Apr.
 & Reopt.
 & Apr.
 & Reopt.\\
 
\midrule
         0.8 / 0.0 / 0.2 & 6.74 & 6.59 & 10.92 & 10.61 & 21.83 & 21.47 & 51.38 & 50.61 \\ 
         0.8 / 0.2 / 0.0 & \textbf{6.31} & \textbf{6.12} & \textbf{10.73} & \textbf{10.41} & \textbf{21.27} & \textbf{20.63} & \textbf{50.45} & \textbf{48.93} \\ 
         0.6 / 0.2 / 0.2 & 6.37 & 6.18 & 10.83 & 10.50 & 21.46 & 20.82 & 50.90 & 49.37 \\ 
         0.4 / 0.3 / 0.3 & 6.83 & 6.62 & 11.60 & 11.25 & 22.99 & 22.30 & 54.54 & 52.90 \\  
\bottomrule

        \end{tabular}
    }
% \vskip -8pt
\centering
\caption{The results derived from two customer positioning approaches on the SVRP test dataset.}
\label{customer_positions}
    \scalebox{0.65}{
        \begin{tabular}{c cc cc cc cc} 
\\
\toprule

 & \multicolumn{2}{c}{{{10}}}
 & \multicolumn{2}{c}{{{20}}}
 & \multicolumn{2}{c}{{{50}}}
 & \multicolumn{2}{c}{{{100}}}\\

 \cmidrule[0.4pt](lr{0.125em}){2-3}%
 \cmidrule[0.4pt](lr{0.125em}){4-5}%
 \cmidrule[0.4pt](lr{0.125em}){6-7}%
 \cmidrule[0.4pt](lr{0.125em}){8-9}%

  Positioning
 & Apr.
 & Reopt.
 & Apr.
 & Reopt.
 & Apr.
 & Reopt.
 & Apr.
 & Reopt.\\
 
\midrule
         Flexible & 7.05 & 6.84 & 11.99 & 11.63 & 23.76 & 23.05 & 56.36 & 54.66 \\ 
         Fixed & \textbf{6.37} & \textbf{6.18} & \textbf{10.83} & \textbf{10.50} & \textbf{21.46} & \textbf{20.82} & \textbf{50.90} & \textbf{49.37} \\ 
         \bottomrule
        \end{tabular}

    }
    \vskip -8pt
\end{table}
In the next experiment, we scrutinized the impact of fixed customer positions versus flexible customer positions. The outcomes of this experiment are depicted in Table \ref{customer_positions}. Evidently, the results favor fixed customer positions due to the less intricate nature of the problem in this scenario. However, the primary objective of this experiment was to assess the model's capacity to learn in both settings. The results affirm that the model can effectively optimize routes in both positioning approaches, with superior performance observed in the case of fixed customer positions.

\begin{figure}%[!ht]
  \centering
\includegraphics[width=.6\linewidth]{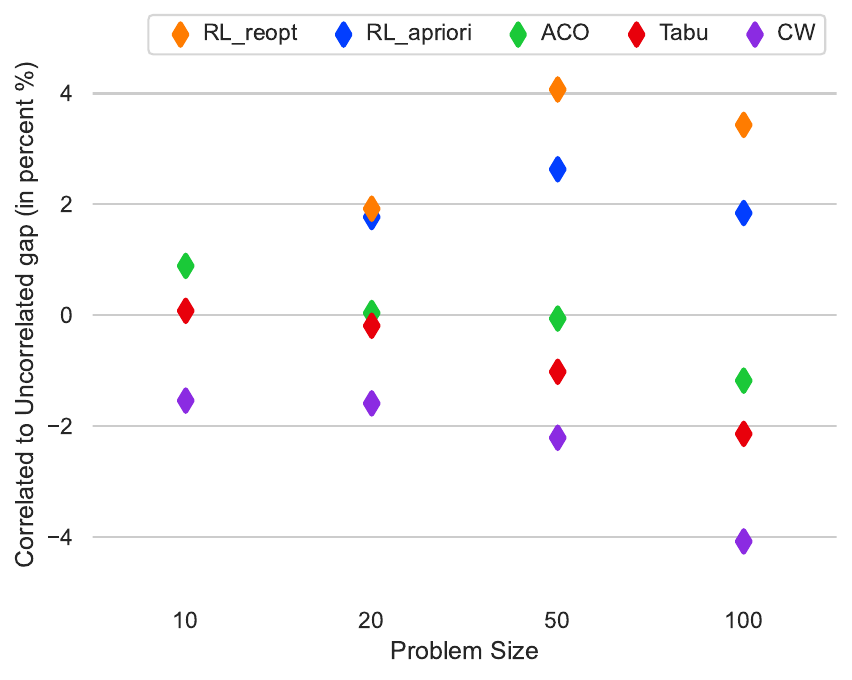} 
  \caption{Percentage difference in travel costs between scenarios with correlated variables (setting $\mathrm{A},\mathrm{B},\Gamma=0.8,0.2,0.0$) and uncorrelated variables (setting $\mathrm{A},\mathrm{B},\Gamma=0.8,0.0,0.2$).}
\label{fig:main_plot}
\vskip -18pt
\end{figure}

In the subsequent experiment, we assessed the capacity of both baselines and the RL model to discern correlations among stochastic variables. This correlation was introduced by maintaining identical values of external variables among customers for both demand and travel costs. The experiment aimed to examine the performance gap between correlated and uncorrelated settings. The outcomes are illustrated in Figure \ref{fig:main_plot}, showcasing that CW and Tabu Search methods are unable to leverage the correlation among stochastic variables. The ACO algorithm exhibits improved performance in correlated environments with smaller-sized problems; however, as the problem size increases, its performance diminishes. Notably, the RL model demonstrates a noteworthy ability to discern and utilize correlations among stochastic variables, resulting in enhanced routing optimization.
\begin{table}
\centering
\caption{The results derived from different fill rates on the SVRP test dataset.}
\label{fill_rate_table}
    \scalebox{0.7}{
        \begin{tabular}{c cc cc cc cc} 
\\
\toprule

 & \multicolumn{2}{c}{{{10}}}
 & \multicolumn{2}{c}{{{20}}}
 & \multicolumn{2}{c}{{{50}}}
 & \multicolumn{2}{c}{{{100}}}\\

 \cmidrule[0.4pt](lr{0.125em}){2-3}%
 \cmidrule[0.4pt](lr{0.125em}){4-5}%
 \cmidrule[0.4pt](lr{0.125em}){6-7}%
 \cmidrule[0.4pt](lr{0.125em}){8-9}%
 
 Rate 
 & Apr.
 & Reopt.
 & Apr.
 & Reopt.
 & Apr.
 & Reopt.
 & Apr.
 & Reopt.\\
 
\midrule
         0.1 & 7.05 & 6.84 & 11.99 & 11.63 & 23.76 & 23.05 & 56.36 & 54.66 \\ 
         0.5 & 6.37 & 6.18 & 10.83 & 10.50 & 21.46 & 20.82 & 50.90 & 49.37 \\ 
         0.9 & \textbf{5.97} & \textbf{5.79} & \textbf{10.15} & \textbf{9.85} & \textbf{20.12} & \textbf{19.52} & \textbf{47.72} & \textbf{46.29} \\ 
\bottomrule
        \end{tabular}
    }
\caption{The results obtained from various delivery types on the SVRP test dataset.}
\centering
\label{supply_type_table}
    \scalebox{0.7}{
        \begin{tabular}{c cc cc cc cc} 
\\
\toprule

 & \multicolumn{2}{c}{{{10}}}
 & \multicolumn{2}{c}{{{20}}}
 & \multicolumn{2}{c}{{{50}}}
 & \multicolumn{2}{c}{{{100}}}\\

 \cmidrule[0.4pt](lr{0.125em}){2-3}%
 \cmidrule[0.4pt](lr{0.125em}){4-5}%
 \cmidrule[0.4pt](lr{0.125em}){6-7}%
 \cmidrule[0.4pt](lr{0.125em}){8-9}%
 
 Delivery
 & Apr.
 & Reopt.
 & Apr.
 & Reopt.
 & Apr.
 & Reopt.
 & Apr.
 & Reopt.\\
 
\midrule

         Full & 7.00 & 6.79 & 11.89 & 11.53 & 23.57 & 22.86 & 55.90 & 54.22 \\ 
         Partial & \textbf{6.37} & \textbf{6.18} & \textbf{10.83} & \textbf{10.50} & \textbf{21.46} & \textbf{20.82} & \textbf{50.90} & \textbf{49.37} \\ 
\bottomrule
        \end{tabular}
    }
\caption{The results obtained with different numbers of vehicles on the SVRP test dataset.}
\centering
\label{mutiple_vehicles}

    \scalebox{0.7}{
        \begin{tabular}{c cc cc cc cc} 
\\
\toprule

 & \multicolumn{2}{c}{{{10}}}
 & \multicolumn{2}{c}{{{20}}}
 & \multicolumn{2}{c}{{{50}}}
 & \multicolumn{2}{c}{{{100}}}\\

 \cmidrule[0.4pt](lr{0.125em}){2-3}%
 \cmidrule[0.4pt](lr{0.125em}){4-5}%
 \cmidrule[0.4pt](lr{0.125em}){6-7}%
 \cmidrule[0.4pt](lr{0.125em}){8-9}%
 
 Vehicles
 & Apr.
 & Reopt.
 & Apr.
 & Reopt.
 & Apr.
 & Reopt.
 & Apr.
 & Reopt.\\
 \midrule
     1 & 6.37 & 6.18 & 10.83 & 10.50 & 21.46 & 20.82 & 50.90 & 49.37 \\ 
     2 & 6.23 & 6.05 & 10.60 & 10.28 & 21.00 & 20.37 & 49.81 & 48.32 \\ 
     3 & 6.19 & 6.01 & 10.53 & 10.21 & 20.87 & 20.24 & 49.49 & 48.01 \\ 
     5 & \textbf{6.17} & \textbf{5.99} & \textbf{10.49} & \textbf{10.18} & \textbf{20.79} & \textbf{20.17} & \textbf{49.31} & \textbf{47.83} \\ 
     \bottomrule
        \end{tabular}
    }
\vskip -18pt
\end{table}
We assessed the model's performance under varying fill rate conditions, as depicted in Table \ref{fill_rate_table}. As expected, an increase in the fill rate correlates with improved performance, as larger vehicle capacities allow for the utilization of fewer routes, subsequently reducing the total travel cost. This experiment aimed to investigate the model's adaptability to different fill rate scenarios and ascertain its capability to learn effective policies across various vehicle maximum capacity settings. The results affirm the model's robustness and its capacity to learn optimal routing strategies within specified fill rate intervals. 

Continuing with our exploration, the subsequent experiment delved into examining the model's robustness when confronted with partial versus full delivery types, with results summarized in Table \ref{supply_type_table}. In both settings, the model demonstrated its learning capability, exhibiting superior performance in the case of partial delivery.
{\bf Multi-Vehicle Scenarios.} In all preceding experiments, we employed a single vehicle for simplicity, primarily to assess the base model's robustness across diverse environments. Subsequently, we investigated the model's capacity to learn effective policies in scenarios involving multiple vehicles. The introduction of a larger fleet implies a more expansive action space, offering the potential for enhanced routing strategies. However, due to the exponential growth of the action space, we sought to examine the model's aptitude for learning in such intricate settings. The results, presented in Table \ref{mutiple_vehicles}, indicate that the model indeed leverages the fleet of vehicles to enhance performance with each additional vehicle. Nevertheless, the observed improvement diminishes, suggesting a point of diminishing returns. We posit that there exists an optimal number of vehicles beyond which further additions do not yield appreciable performance gains, warranting exploration in future research.
\section{Conclusion}
This study introduces a formulation of the stochastic vehicle routing problem incorporating time windows, stochastic demands, and travel costs. Bridging a gap in the literature, we propose the an end-to-end framework to address these SVRP challenges. To validate the efficacy and versatility of the model, comprehensive experiments were conducted, evaluating its performance across diverse environmental settings. The experiments encompass variations in inference approaches, delivery types, stochasticity levels, fleet sizes, and customer counts, demonstrating that the model is robust in a range of SVRP configurations.

Additionally, the model design incorporates external variables that closely mirrors real-world scenarios encountered by industry practitioners. Demonstrating a noteworthy 1.73\% improvement against the state-of-the-art ACO metaheuristic, our model exhibited enhanced efficiency in minimizing the total travel cost. This outcome underscores the potential for further research avenues exploring the application of RL in the SVRP domain. Future research trajectories, both in industry and academia, may delve into more intricate architectures or advanced RL training algorithms aimed at mitigating logistic expenses and alleviating associated environmental impacts. Consequently, our model may serve as a versatile starting point for diverse research in SVRP.

\section{Impact Statement}

This research paper on the application of reinforcement learning to the Vehicle Routing Problem holds several significant implications, both within the academic community and in practical, real-world scenarios. By integrating RL into VRP solutions, this research contributes to the evolution of computational logistics. It demonstrates the potential for more adaptive, efficient, and dynamic routing solutions, which can significantly enhance logistical operations. The findings have the potential to revolutionize industry practices by optimizing route planning. This could lead to substantial cost savings, improved delivery times, and greater overall efficiency in transportation and distribution sectors. Optimized routing not only has economic benefits but also environmental ones. More efficient routes mean reduced fuel consumption and lower carbon emissions, contributing to environmentally sustainable business practices. Improved logistics can have a broad societal impact, such as more timely deliveries, reduced traffic congestion, and potential for increased accessibility of goods and services, especially in remote or underserved areas. While enhancing efficiency, the application of RL in VRP also raises important ethical and employment considerations. The technology should be applied in a manner that considers its impact on job markets and the ethical use of artificial intelligence in decision-making processes.

\clearpage
\bibliography{main}

\begin{thebibliography}{40}
\providecommand{\natexlab}[1]{#1}
\providecommand{\url}[1]{\texttt{#1}}
\expandafter\ifx\csname urlstyle\endcsname\relax
  \providecommand{\doi}[1]{doi: #1}\else
  \providecommand{\doi}{doi: \begingroup \urlstyle{rm}\Url}\fi

\bibitem[Andrychowicz et~al.(2020)Andrychowicz, Baker, Chociej, J{\'o}zefowicz, McGrew, Pachocki, Petron, Plappert, Powell, Ray, Schneider, Sidor, Tobin, Welinder, Weng, and Zaremba]{Andrychowicz2020LearningDI}
Andrychowicz, M., Baker, B., Chociej, M., J{\'o}zefowicz, R., McGrew, B., Pachocki, J.~W., Petron, A., Plappert, M., Powell, G., Ray, A., Schneider, J., Sidor, S., Tobin, J., Welinder, P., Weng, L., and Zaremba, W.
\newblock Learning dexterous in-hand manipulation.
\newblock \emph{The International Journal of Robotics Research}, 39:\penalty0 20 -- 3, 2020.

\bibitem[Bello et~al.(2016)Bello, Pham, Le, Norouzi, and Bengio]{bello2016neural}
Bello, I., Pham, H., Le, Q.~V., Norouzi, M., and Bengio, S.
\newblock Neural combinatorial optimization with reinforcement learning.
\newblock \emph{arXiv preprint arXiv:1611.09940}, 2016.

\bibitem[Bello et~al.(2017)Bello, Pham, Le, Norouzi, and Bengio]{Bello2017NeuralCO}
Bello, I., Pham, H., Le, Q.~V., Norouzi, M., and Bengio, S.
\newblock Neural combinatorial optimization with reinforcement learning.
\newblock \emph{ArXiv}, abs/1611.09940, 2017.

\bibitem[Bomboi et~al.(2021)Bomboi, Buchheim, and Pruente]{Bomboi2021OnTS}
Bomboi, F., Buchheim, C., and Pruente, J.
\newblock On the stochastic vehicle routing problem with time windows, correlated travel times, and time dependency.
\newblock \emph{4OR}, pp.\  1--23, 2021.

\bibitem[Christiansen \& Lysgaard(2007)Christiansen and Lysgaard]{Christiansen2007ABA}
Christiansen, C.~H. and Lysgaard, J.
\newblock A branch-and-price algorithm for the capacitated vehicle routing problem with stochastic demands.
\newblock \emph{Oper. Res. Lett.}, 35:\penalty0 773--781, 2007.

\bibitem[Cordeau et~al.(2007)Cordeau, Laporte, Savelsbergh, and Vigo]{inbook}
Cordeau, J.-F., Laporte, G., Savelsbergh, M.~W., and Vigo, D.
\newblock Vehicle routing.
\newblock \emph{Handbooks in operations research and management science}, 14:\penalty0 367--428, 2007.

\bibitem[Dantzig \& Ramser(1959)Dantzig and Ramser]{Dantzig1959TheTD}
Dantzig, G.~B. and Ramser, J.~H.
\newblock The truck dispatching problem.
\newblock \emph{Management Science}, 6:\penalty0 80--91, 1959.

\bibitem[{De La Vega} et~al.(2023){De La Vega}, Gendreau, Morabito, Munari, and Ordóñez]{DELAVEGA2023676}
{De La Vega}, J., Gendreau, M., Morabito, R., Munari, P., and Ordóñez, F.
\newblock An integer l-shaped algorithm for the vehicle routing problem with time windows and stochastic demands.
\newblock \emph{European Journal of Operational Research}, 308\penalty0 (2):\penalty0 676--695, 2023.
\newblock ISSN 0377-2217.
\newblock \doi{https://doi.org/10.1016/j.ejor.2022.11.040}.
\newblock URL \url{https://www.sciencedirect.com/science/article/pii/S0377221722008979}.

\bibitem[Fukasawa \& Gunter(2023)Fukasawa and Gunter]{FUKASAWA202311}
Fukasawa, R. and Gunter, J.
\newblock The complexity of branch-and-price algorithms for the capacitated vehicle routing problem with stochastic demands.
\newblock \emph{Operations Research Letters}, 51\penalty0 (1):\penalty0 11--16, 2023.
\newblock ISSN 0167-6377.
\newblock \doi{https://doi.org/10.1016/j.orl.2022.11.005}.
\newblock URL \url{https://www.sciencedirect.com/science/article/pii/S0167637722001468}.

\bibitem[Gauvin et~al.(2014)Gauvin, Desaulniers, and Gendreau]{Gauvin2014ABA}
Gauvin, C., Desaulniers, G., and Gendreau, M.
\newblock A branch-cut-and-price algorithm for the vehicle routing problem with stochastic demands.
\newblock \emph{Comput. Oper. Res.}, 50:\penalty0 141--153, 2014.

\bibitem[Gendreau et~al.(1996)Gendreau, Laporte, and S{\'e}guin]{GENDREAU19963}
Gendreau, M., Laporte, G., and S{\'e}guin, R.
\newblock Stochastic vehicle routing.
\newblock \emph{European Journal of Operational Research}, 88\penalty0 (1):\penalty0 3--12, 1996.

\bibitem[Goel et~al.(2019)Goel, Maini, and Bansal]{Goel2019VehicleRP}
Goel, R.~K., Maini, R., and Bansal, S.
\newblock Vehicle routing problem with time windows having stochastic customers demands and stochastic service times: Modelling and solution.
\newblock \emph{J. Comput. Sci.}, 34:\penalty0 1--10, 2019.

\bibitem[Google(2023)]{Google}
Google.
\newblock Or-tools, 2023.
\newblock URL \url{https://developers.google.com/optimization}.

\bibitem[Helsgaun(2017)]{Helsgaun2017AnEO}
Helsgaun, K.
\newblock An extension of the lin-kernighan-helsgaun tsp solver for constrained traveling salesman and vehicle routing problems: Technical report.
\newblock 2017.

\bibitem[Hildebrandt et~al.(2023)Hildebrandt, Thomas, and Ulmer]{HILDEBRANDT2023106071}
Hildebrandt, F.~D., Thomas, B.~W., and Ulmer, M.~W.
\newblock Opportunities for reinforcement learning in stochastic dynamic vehicle routing.
\newblock \emph{Computers \& Operations Research}, 150:\penalty0 106071, 2023.
\newblock ISSN 0305-0548.
\newblock \doi{https://doi.org/10.1016/j.cor.2022.106071}.
\newblock URL \url{https://www.sciencedirect.com/science/article/pii/S030505482200301X}.

\bibitem[Iklassov et~al.(2023{\natexlab{a}})Iklassov, Medvedev, de~Retana, and Tak{\'{a}}c]{IklassovMR023}
Iklassov, Z., Medvedev, D., de~Retana, R. S.~O., and Tak{\'{a}}c, M.
\newblock On the study of curriculum learning for inferring dispatching policies on the job shop scheduling.
\newblock In \emph{Proceedings of the Thirty-Second International Joint Conference on Artificial Intelligence, {IJCAI} 2023, 19th-25th August 2023, Macao, SAR, China}, pp.\  5350--5358. ijcai.org, 2023{\natexlab{a}}.
\newblock \doi{10.24963/IJCAI.2023/594}.
\newblock URL \url{https://doi.org/10.24963/ijcai.2023/594}.

\bibitem[Iklassov et~al.(2023{\natexlab{b}})Iklassov, Sobirov, Solozabal, and Takac]{iklassov2023reinforcement}
Iklassov, Z., Sobirov, I., Solozabal, R., and Takac, M.
\newblock Reinforcement learning for solving stochastic vehicle routing problem, 2023{\natexlab{b}}.

\bibitem[Iklassov et~al.(2023{\natexlab{c}})Iklassov, Sobirov, Solozabal, and Tak{\'{a}}c]{iklassovSST23}
Iklassov, Z., Sobirov, I., Solozabal, R., and Tak{\'{a}}c, M.
\newblock Reinforcement learning approach to stochastic vehicle routing problem with correlated demands.
\newblock \emph{{IEEE} Access}, 11:\penalty0 87958--87969, 2023{\natexlab{c}}.
\newblock \doi{10.1109/ACCESS.2023.3306076}.
\newblock URL \url{https://doi.org/10.1109/ACCESS.2023.3306076}.

\bibitem[Jin et~al.(2023)Jin, Cui, Bai, and Qu]{JIN2023}
Jin, J., Cui, T., Bai, R., and Qu, R.
\newblock Container port truck dispatching optimization using real2sim based deep reinforcement learning.
\newblock \emph{European Journal of Operational Research}, 2023.
\newblock ISSN 0377-2217.
\newblock \doi{https://doi.org/10.1016/j.ejor.2023.11.038}.
\newblock URL \url{https://www.sciencedirect.com/science/article/pii/S0377221723008792}.

\bibitem[Joshi et~al.(2019)Joshi, Laurent, and Bresson]{joshi2019efficient}
Joshi, C.~K., Laurent, T., and Bresson, X.
\newblock An efficient graph convolutional network technique for the travelling salesman problem.
\newblock \emph{arXiv preprint arXiv:1906.01227}, 2019.

\bibitem[Kool et~al.(2018)Kool, Van~Hoof, and Welling]{kool2018attention}
Kool, W., Van~Hoof, H., and Welling, M.
\newblock Attention, learn to solve routing problems!
\newblock \emph{arXiv preprint arXiv:1803.08475}, 2018.

\bibitem[Laporte(2009)]{Laporte2009FiftyYO}
Laporte, G.
\newblock Fifty years of vehicle routing.
\newblock \emph{Transp. Sci.}, 43:\penalty0 408--416, 2009.

\bibitem[Laporte \& Louveaux(1993)Laporte and Louveaux]{Laporte1993TheIL}
Laporte, G. and Louveaux, F.~V.
\newblock The integer l-shaped method for stochastic integer programs with complete recourse.
\newblock \emph{Oper. Res. Lett.}, 13:\penalty0 133--142, 1993.

\bibitem[Li \& Li(2020)Li and Li]{Li2020AnIT}
Li, G. and Li, J.
\newblock An improved tabu search algorithm for the stochastic vehicle routing problem with soft time windows.
\newblock \emph{IEEE Access}, 8:\penalty0 158115--158124, 2020.

\bibitem[Li et~al.(2021)Li, Yan, and Wu]{Wang2021AlphaTLT}
Li, S., Yan, Z., and Wu, C.
\newblock Learning to delegate for large-scale vehicle routing.
\newblock \emph{Neural Information Processing Systems}, 2021.
\newblock URL \url{https://api.semanticscholar.org/CorpusID:235790288}.

\bibitem[Louveaux(1998)]{Louveaux1998AnIT}
Louveaux, F.~V.
\newblock An introduction to stochastic transportation models.
\newblock 1998.

\bibitem[Lu et~al.(2020)Lu, Zhang, and Yang]{Lu2020ALI}
Lu, H., Zhang, X., and Yang, S.
\newblock A learning-based iterative method for solving vehicle routing problems.
\newblock In \emph{ICLR}, 2020.

\bibitem[Nazari et~al.(2018)Nazari, Oroojlooy, Snyder, and Tak{\'a}{\v{c}}]{nazari2018reinforcement}
Nazari, M., Oroojlooy, A., Snyder, L.~V., and Tak{\'a}{\v{c}}, M.
\newblock Reinforcement learning for solving the vehicle routing problem.
\newblock In \emph{Conference on Neural Information Processing Systems, NeurIPS 2018}, 2018.

\bibitem[Oroojlooyjadid et~al.(2022)Oroojlooyjadid, Nazari, Snyder, and Tak{\'a}c]{Oroojlooyjadid2022ADQ}
Oroojlooyjadid, A., Nazari, M., Snyder, L.~V., and Tak{\'a}c, M.
\newblock A deep q-network for the beer game: Deep reinforcement learning for inventory optimization.
\newblock \emph{Manuf. Serv. Oper. Manag.}, 24:\penalty0 285--304, 2022.

\bibitem[Pichpibul \& Kawtummachai(2013)Pichpibul and Kawtummachai]{Pichpibul2013AHA}
Pichpibul, T. and Kawtummachai, R.
\newblock A heuristic approach based on clarke-wright algorithm for open vehicle routing problem.
\newblock \emph{The Scientific World Journal}, 2013, 2013.

\bibitem[Rajabi-Bahaabadi et~al.(2021)Rajabi-Bahaabadi, Shariat, Babaei, and Vigo]{Bahaabadi2021}
Rajabi-Bahaabadi, M., Shariat, A., Babaei, M., and Vigo, D.
\newblock Reliable vehicle routing problem in stochastic networks with correlated travel times.
\newblock \emph{Operational Research}, 03 2021.
\newblock \doi{10.1007/s12351-019-00452-w}.

\bibitem[Silver et~al.(2017)Silver, Schrittwieser, Simonyan, Antonoglou, Huang, Guez, Hubert, Baker, Lai, Bolton, et~al.]{silver2017mastering}
Silver, D., Schrittwieser, J., Simonyan, K., Antonoglou, I., Huang, A., Guez, A., Hubert, T., Baker, L., Lai, M., Bolton, A., et~al.
\newblock Mastering the game of go without human knowledge.
\newblock \emph{nature}, 550\penalty0 (7676):\penalty0 354--359, 2017.

\bibitem[Sutton et~al.(1999)Sutton, McAllester, Singh, and Mansour]{sutton1999policy}
Sutton, R.~S., McAllester, D., Singh, S., and Mansour, Y.
\newblock Policy gradient methods for reinforcement learning with function approximation.
\newblock \emph{Advances in neural information processing systems}, 12, 1999.

\bibitem[Toth \& Vigo(2014)Toth and Vigo]{Toth2014VehicleRP}
Toth, P. and Vigo, D.
\newblock Vehicle routing: Problems, methods, and applications, second edition.
\newblock 2014.

\bibitem[Tricks(2018)]{Tricks2018AND}
Tricks, L.~O.
\newblock A new dog learns old tricks : Rl finds classic optimization algorithms.
\newblock 2018.

\bibitem[Wang et~al.(2021)Wang, Hu, Wang, Xu, Ma, Yang, Liu, and Wang]{wang2021dynamic}
Wang, L., Hu, X., Wang, Y., Xu, S., Ma, S., Yang, K., Liu, Z., and Wang, W.
\newblock Dynamic job-shop scheduling in smart manufacturing using deep reinforcement learning.
\newblock \emph{Computer Networks}, 190:\penalty0 107969, 2021.

\bibitem[Williams(1992)]{williams1992simple}
Williams, R.~J.
\newblock Simple statistical gradient-following algorithms for connectionist reinforcement learning.
\newblock \emph{Machine learning}, 8\penalty0 (3):\penalty0 229--256, 1992.

\bibitem[Zhang et~al.(2023{\natexlab{a}})Zhang, Luo, Florio, and {Van Woensel}]{ZHANG2023596}
Zhang, J., Luo, K., Florio, A.~M., and {Van Woensel}, T.
\newblock Solving large-scale dynamic vehicle routing problems with stochastic requests.
\newblock \emph{European Journal of Operational Research}, 306\penalty0 (2):\penalty0 596--614, 2023{\natexlab{a}}.
\newblock ISSN 0377-2217.
\newblock \doi{https://doi.org/10.1016/j.ejor.2022.07.015}.
\newblock URL \url{https://www.sciencedirect.com/science/article/pii/S0377221722005677}.

\bibitem[Zhang et~al.(2023{\natexlab{b}})Zhang, Ji, and Yu]{su15021741}
Zhang, Z., Ji, B., and Yu, S.~S.
\newblock An adaptive tabu search algorithm for solving the two-dimensional loading constrained vehicle routing problem with stochastic customers.
\newblock \emph{Sustainability}, 15\penalty0 (2), 2023{\natexlab{b}}.
\newblock ISSN 2071-1050.
\newblock \doi{10.3390/su15021741}.
\newblock URL \url{https://www.mdpi.com/2071-1050/15/2/1741}.

\bibitem[Zhou et~al.(2023)Zhou, Ma, Douge, Chew, and Lee]{ZHOU2023109443}
Zhou, C., Ma, J., Douge, L., Chew, E.~P., and Lee, L.~H.
\newblock Reinforcement learning-based approach for dynamic vehicle routing problem with stochastic demand.
\newblock \emph{Computers \& Industrial Engineering}, 182:\penalty0 109443, 2023.
\newblock ISSN 0360-8352.
\newblock \doi{https://doi.org/10.1016/j.cie.2023.109443}.
\newblock URL \url{https://www.sciencedirect.com/science/article/pii/S0360835223004679}.

\end{thebibliography}
\bibliographystyle{icml2024}
%%%%%%%%%%%%%%%%%%%%%%%%%%%%%%%%%%%%%%%%%%%%%%%%%%%%%%%%%%%%%%%%%%%%%%%%%%%%%%%
%%%%%%%%%%%%%%%%%%%%%%%%%%%%%%%%%%%%%%%%%%%%%%%%%%%%%%%%%%%%%%%%%%%%%%%%%%%%%%%
% APPENDIX
%%%%%%%%%%%%%%%%%%%%%%%%%%%%%%%%%%%%%%%%%%%%%%%%%%%%%%%%%%%%%%%%%%%%%%%%%%%%%%%
%%%%%%%%%%%%%%%%%%%%%%%%%%%%%%%%%%%%%%%%%%%%%%%%%%%%%%%%%%%%%%%%%%%%%%%%%%%%%%%
\newpage
\appendix
\onecolumn

\section{Classical Formulation}
\label{formulation_apdx}

The stochasticity in the SVRP can be attributed to the following three bases, according to~\cite{Toth2014VehicleRP}:
\begin{enumerate}[noitemsep]
  \item \textbf{stochastic demands} (VRPSD), i.e., uncertain customer demands,
  \item \textbf{stochastic customers} (VRPSC), i.e., uncertain presence/absence of customers,
  \item \textbf{stochastic travel costs} (VRPSTT), i.e., uncertain travel costs for various routes. 
\end{enumerate}

The deterministic character of VRP falls short of reaching an optimal solution for the SVRP, as noted in the study~\cite{Louveaux1998AnIT}. The following formulation is a mathematical representation of SVRP used in the classical research literature.

The objective of the problem is to
\begin{align}
\text{minimize}\qquad &\textstyle{\sum}_{i, j \in C} c_{i j} x_{i j}+\mathscr{R}(x), \label{eq:app_1} \\
\text{subject to }\qquad & \textstyle{\sum}_{j=1}^{n} x_{0 j}=2|K|, \label{eq:app_2} \\
& \textstyle{\sum}_{i<k} x_{i k}+\textstyle{\sum}_{j>k} x_{k j}=2 \quad \forall k \in N, \label{eq:app_3} \\
& \textstyle{\sum}_{i, j \in C} x_{i j} \leq|C|-[\sum_{i \in C} \mathbb{E} [d_{i}] / Q], \label{eq:app_4} \\
& 0 \leq x_{i j} \leq 1 \quad \forall i, j \in C, \label{eq:app_5} \\
& 0 \leq x_{0 j} \leq 2 \quad \forall j \in N, \label{eq:app_6} \\
& x=\left(x_{i j}\right) \text { integer } \quad \forall i, j \in N. \label{eq:app_7}
\end{align}

% where,
% \\
% $N$ : set of customers and depot,
% \\
% $C$ : customers set,
% \\
% $c_{i j}$ : stochastic travel cost between nodes $i$ and $j$,
% \\
% $d_{i}$ : stochastic demand of customer $i$,
% \\
% $K$ : number of vehicles,
% \\
% $Q$ : maximum capacity of each vehicle,
% \\
% $x_{i j}:$ : Binary variable indicating whether the route utilizes the edge $(i, j)$.

The problem is defined using the following notations: $C$ represents the customer set, $c_{i j}$ stands for the variable travel cost between nodes $i$ and $j$, $x_{i j}$ denotes a binary variable that identifies if $(i, j)$ is used in the route. This formulation of the SVRP is formed in a complete, undirected graph $G=(N, X)$, where $N=\{0,1, \ldots, n\}$ signifies the depot node (at the node 0) and the customer nodes. The depot is the root point for a fleet of $K$ vehicles, each loaded with $Q$ initial capacity. For each consumer node $i \in C=N \backslash{0}$, there is a stochastic demand $\xi_{i}$ and its $s^{th}$ realization denoted as $\mu_{is}$. A set of arcs, $X={(i, j): i, j \in N, i<j}$, represent connections between nodes, with $c_{ij}$ defining the travel cost for an arc $(i, j) \in X$. The action of traversing an arc $(i, j)$ is represented by $x_{i j}$, where $x_{i j}$=1 means the arc is included in the route, 0 otherwise. The algorithm also comprises a recourse cost $\mathscr{R}(x)$ for situations where a vehicle cannot satisfy the demand of a customer due to limited capacity, accounting for the extra cost of going back to the depot for a reload.

The formulation aims to minimize the overall traversal cost, as expressed in Equation~\ref{eq:app_1}, which is the summation of costs associated with each arc in the route. Various constraints are established to ensure that (i) each vehicle commences and concludes its journey at a designated depot (Equation \ref{eq:app_2}), (ii) each customer is visited exactly once (Equation \ref{eq:app_3}), (iii) the maximum load of each vehicle accommodates the expected demand (Equation \ref{eq:app_4}), and (iv) the decision variable $x_{ij}$ is an integer for every arc (Equations \ref{eq:app_5},\ref{eq:app_6},\ref{eq:app_7}). These constraints are implemented to derive a solution that adheres to the inherent constraints of the problem.

In instances where a vehicle is unable to satisfy a customer's demand due to insufficient load, the recourse cost $\mathscr{R}(x)$ specified in Equation \ref{eq:app_1} addresses this situation. Specifically, this term captures the cost associated with traveling to the depot and returning for refilling to fulfill the customer's demand. The mathematical representation of the incurred cost is expressed as follows:

\begin{align} 
    \mathscr{R}(x)&=\textstyle{\sum}_{k=1}^{K}\mathscr{R}^{k}(x), \nonumber \\
    \mathscr{R}^{k}(x)&=2\textstyle{\sum}_{j=2}^{t}\textstyle{\sum}_{l=1}^{j-1}P(\textstyle{\sum}_{s=2}^{j-1}\xi_{s}\leq l Q<\textstyle{\sum}_{s=2}^{j}\xi_{s})c_{0 j}.  \nonumber 
\end{align}

the total recourse cost is the summation of the incremental cost incurred by each vehicle ($k$), expressed as $\mathscr{R}^{k}(x)$. This calculation assesses the probability of encountering the $l^{th}$ failure scenario at the $j^{th}$ customer along the route.

\begin{table}
\caption{The performance of the ACO algorithm. The SVRP validation dataset was utilized to assess the efficacy of different hyperparameter settings (pheromone importance / heuristic importance / number of ants). Only best results are depicted. Each column corresponds to a varying number of customers. The outcomes are depicted in terms of the average travel cost across the validation dataset, where diminished values signify superior performance.}

\label{hyperparam_search_table}
\centering
\begin{small}
\begin{sc}
\scalebox{0.9}{
\begin{tabular}{crrrr}
\\
\toprule
    ACO parameters &    10 &    20 &     50 &    100 \\
\midrule
         3 / 4 / 20 & 6.77 & 11.51 & 22.82 & 54.13 \\
         6 / 8 / 24 & 6.73 & 11.44 & 22.67 & 53.77 \\ 
         4 / 9 / 15 & 6.67 & 11.33 & 22.46 & 53.27 \\ 
         3 / 10/ 12 & \textbf{6.48} & \textbf{11.02} & \textbf{21.84} & \textbf{51.81} \\ 

\bottomrule
\end{tabular}
}
\end{sc}
\end{small}
\end{table}

\begin{table} % title of Table
\centering % used for centering table
\caption{Hyperparameters values.}
\begin{tabular}{l@{\hskip 0.2in}c@{\hskip 0.2in}}% centered columns (4 columns)
\toprule %inserts double horizontal lines
\textbf{Hyperparameter} & \textbf{Value} \\ %[0.5ex] % inserts table
%heading
\midrule % inserts single horizontal line
% & \multicolumn{2}{c}{Four Rooms }\\
Customer input dimension	& $(N+19) \times (N+1)$ \\
Vehicle input dimension	& $4 \times K$ \\
Embedding dimension	& $128$ \\
LSTM [layer, dim]	& $[1,128]$\\
Critic [depth,width]	& $[2,128]$ \\
Actor [output]	& $K \times (N+1)$ \\
Activation function	& ReLU \\
Learning rate	& 1E-04 \\
Optimizer	& Adam \\
Batch size 	& $128$	\\
Training iterations & $10000$ \\
\bottomrule
\end{tabular}
\label{tab:hyperparameters}
\end{table}

\newpage

\section{Implementation}

{\bf Critic Network.} The outputs, denoted as $P^{t+1}$, derived from this process are subsequently fed into a Critic network. This network comprises two fully connected layers, each with $D$ and one neuron, and employs the Rectified Linear Unit (ReLU) activation function. The dimensionality of the embeddings denoted as $D$, is defined as $128$.

{\bf ACO Tuning.} Given the sensitivity of the ACO model, we conducted its hyperparameter search. The outcomes of this search are presented in Table~\ref{hyperparam_search_table}. Hyperparameters specified in the respective papers were employed for the other baselines.

{\bf Training Details.} Hyperparameter tuning was conducted using a grid search approach to refine the model. Various configurations were tested, including LSTM and GRU cells, as well as recurrent and convolutional layers for vehicle embeddings. We evaluated different activation functions such as Sigmoid, LeakyRELU, RELU, ELU, and tanh, alongside a range of optimizers including Nesterov, AdaGrad, AdaDelta, and Adam. Additionally, embedding dimensions (64, 128, 256, 512), learning rates, and sizes of the Critic network were examined. The optimal network configuration is detailed in Table \ref{tab:hyperparameters}. We adopt the Xavier initialization method and employ the Adam optimizer with a learning rate set to $10^{-4}$ during the training phase. To mitigate overfitting, a dropout with a probability of 0.1 is applied. The training process is executed on a GPU system, specifically a NVIDIA A100 SXM 40GB GPU and 2x AMD EPYC 7742 CPUs (8 cores) with 256GB RAM. The training duration consists of a total of 10,000 iterations for each problem size. 

% TODO 
    % Describe tab:hyperparameters
    % Explain how hyperparameters were chosen 
    % hyper tuning similar to ACO

\newpage

\section{Time Complexity.}

\begin{table}
\caption{Results for the inference time in seconds on the SVRP test dataset.}
\centering
\label{comp_times}
    \scalebox{0.9}{
        \begin{tabular}{l c c c c} 
\\
\toprule
 Baseline
 & 10
 & 20
 & 50
 & 100\\

\midrule
    Clarke-Wright &	0.005 &	0.016 &	0.058 &	0.185 \\ 
    Tabu Search &	0.964 &	2.936 &	8.672 &	57.12 \\ 
    ACO &	1.230 &	4.291 &	27.64 &	123.7 \\ 
    RL-greedy &	0.061 &	0.121 &	0.191 &	0.394 \\ 
    RL-sampling &	0.064 &	0.125 &	0.232 &	0.405 \\ 
    RL-beam search &	0.072 &	0.178 &	0.293 &	0.416 \\ 
    \bottomrule
        \end{tabular}
    }
\end{table}

In a concluding assessment, we examined the inference times of baseline methods and our proposed model, as detailed in Table \ref{comp_times}. Evidently, the CW heuristic demonstrates the shortest time required to find a solution. Our model exhibits the next most efficient inference timing. Future studies could broaden this analysis to include larger datasets, possibly with thousands of customers, to further evaluate the model's suitability for real-world logistics applications and to balance the benefits in inference speed against the costs associated with training time.

\newpage

\section{Training Curves}

\begin{figure}%[!ht]
\centering
\begin{tabular}{ c @{\hspace{20pt}} c }
  \includegraphics[width=.25\linewidth]{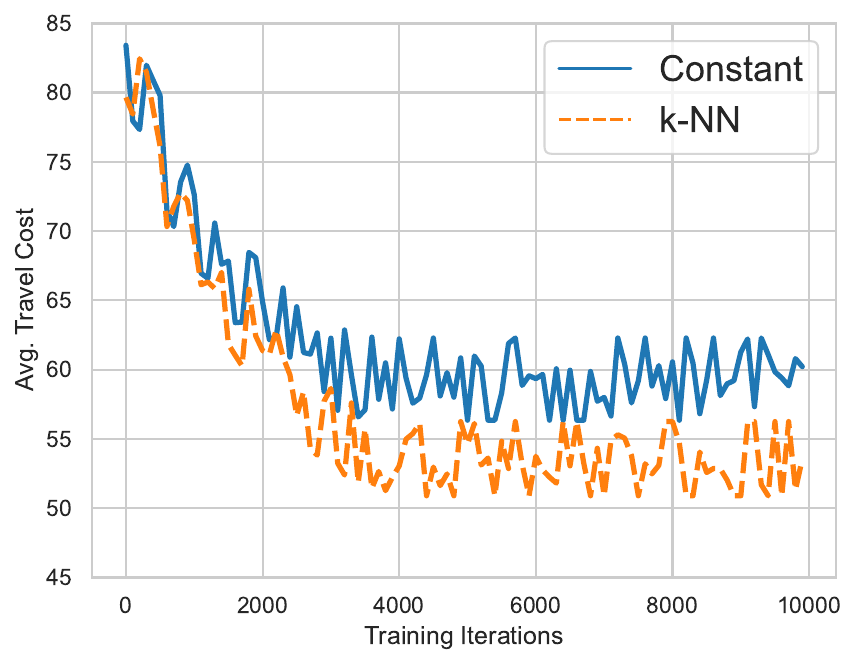} &
  \includegraphics[width=.25\linewidth]{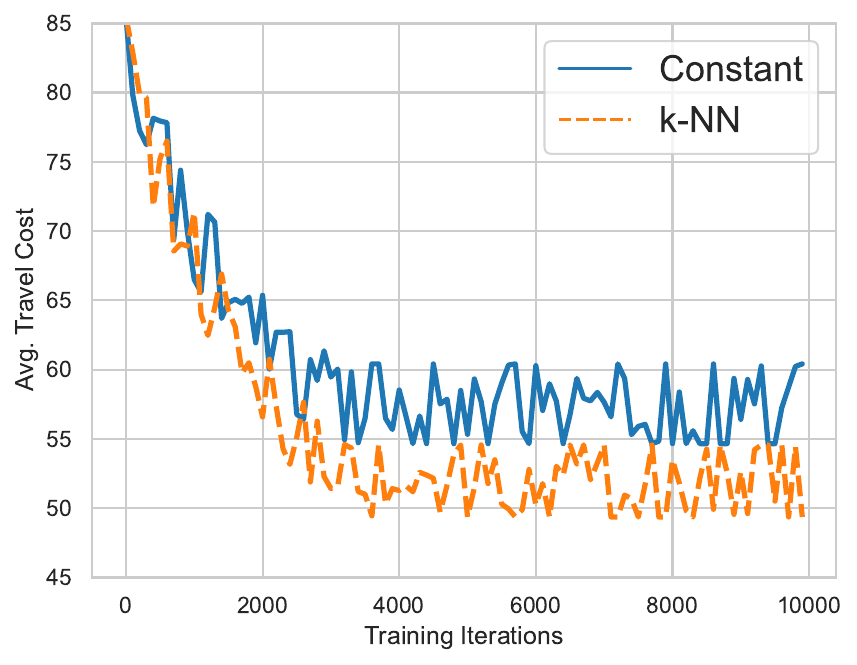} \\
  \small A priori &
  \small Reoptimization
\end{tabular}
\medskip
\caption{The incurred travel cost throughout the training phase, employing two distinct customer position approaches: fixed and flexible.}
\label{fig:position}
\end{figure}

\begin{figure}%[!ht]
\centering
\begin{tabular}{ c @{\hspace{20pt}} c }
  \includegraphics[width=.25\linewidth]{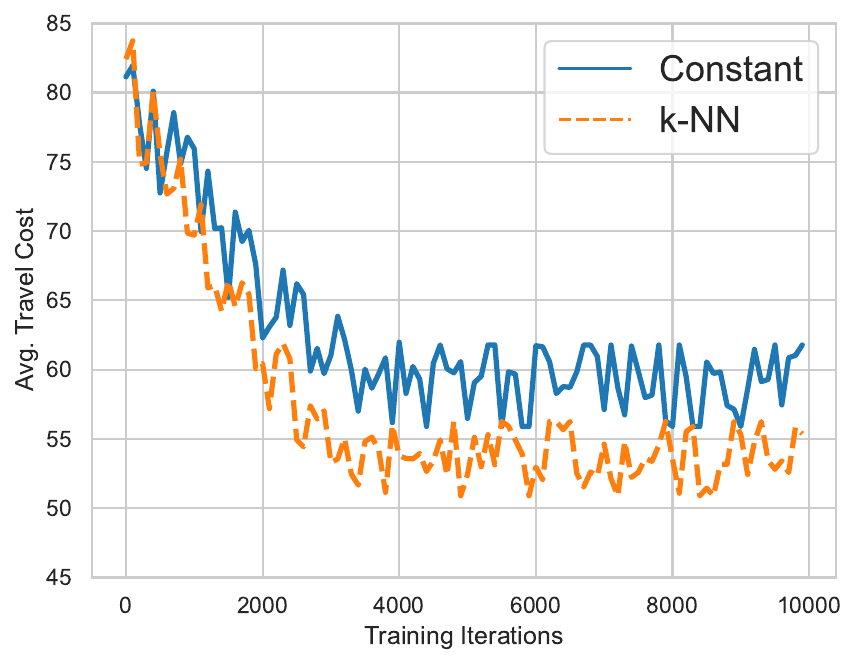} &
  \includegraphics[width=.25\linewidth]{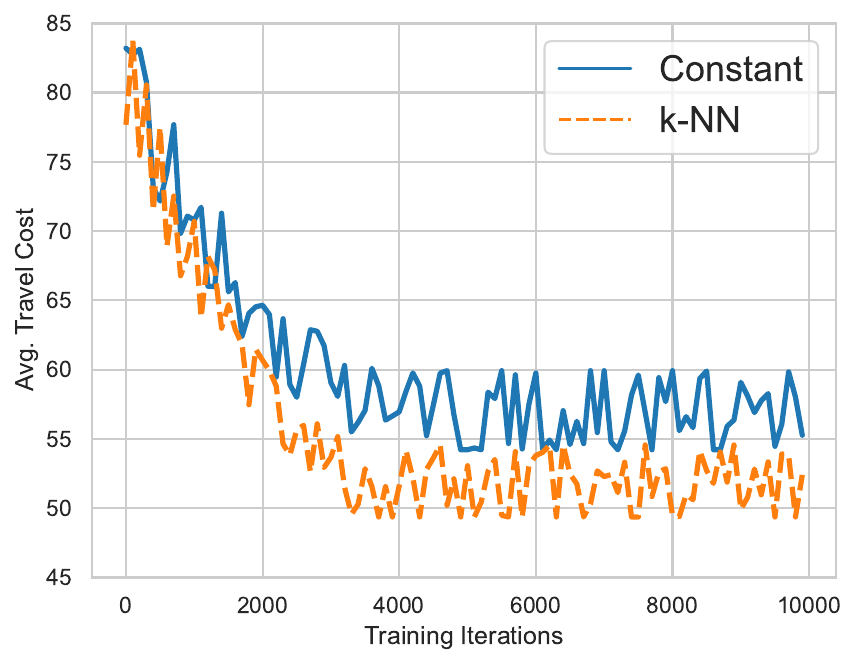} \\
  \small A priori &
  \small Reoptimization
\end{tabular}
\medskip
\caption{The travel cost incurred during the training phase, employing two different delivery approaches: full and partial.}
\label{fig:supply}
\end{figure}

Figure \ref{fig:position} depicts the training curves, providing a comparative analysis of the performance under distinct customer positioning approaches. Additionally, Figure \ref{fig:supply} illustrates the training curves, showcasing the performance variations corresponding to different delivery types.

\section*{Accessibility}
The programming code utilized for the experiments, along with the corresponding software, data and results, have been made publicly available online and can be accessed via the link \footnote{\url{https://github.com/Zangir/SVRP}}

\end{document}